\documentclass[11pt]{article}

\usepackage[preprint]{acl}

\usepackage{times}
\usepackage{latexsym}
\usepackage{multirow}
\usepackage[T1]{fontenc}

\usepackage[utf8]{inputenc}

\usepackage{microtype}

\usepackage{inconsolata}

\usepackage{graphicx}
\usepackage{amsmath}
\usepackage{float} 
\usepackage{enumitem}
\usepackage{titlesec}
\titlespacing*{\section}{0pt}{10pt}{5pt}      
\titlespacing*{\subsection}{0pt}{8pt}{4pt}
\titlespacing*{\subsubsection}{0pt}{6pt}{3pt}
\titlespacing*{\paragraph}{0pt}{3pt}{1em}

\title{Rethinking Scale: Deployment Trade-offs of Small Language Models under Agent Paradigms\thanks{This arXiv version includes Mats Brorsson who was inadvertently omitted from the conference version (ACL industry track, 2026) due to a submission error. All authors have consented to this inclusion.}}

\author{Xinlin Wang \\
  Proximus Luxembourg S.A. \\  18 rue du Puits Romain \\  L-8070 Bertrange, Luxembourg
 \\
  \texttt{xinlin.wang@proximus.lu} \\\And
  Mats Brorsson\\
  University of Luxembourg\\
  6, rue Richard Coudenhove-Kalergi \\
  L-1359 Luxembourg\\
  \texttt{mats.brorrson@uni.lu} \\}

\begin{document}
\maketitle
\begin{abstract}
Despite the impressive capabilities of large language models, their substantial computational costs, latency, and privacy risks hinder their widespread deployment in real-world applications. Small Language Models (SLMs) with fewer than 10 billion parameters present a promising alternative; however, their inherent limitations in knowledge and reasoning curtail their effectiveness. Existing research primarily focuses on enhancing SLMs through scaling laws or fine-tuning strategies while overlooking the potential of using agent paradigms, such as tool use and multi-agent collaboration, to systematically compensate for the inherent weaknesses of small models. 
To address this gap, this paper presents the first large-scale, comprehensive study of <10B open-source models under three paradigms: (1) the base model, (2) a single agent equipped with tools, and (3) a multi-agent system with collaborative capabilities.
Our results show that single-agent systems achieve the best balance between performance and cost, while multi-agent setups add overhead with limited gains. Our findings highlight the importance of agent-centric design for efficient and trustworthy deployment in resource-constrained settings.
\end{abstract}

\section{Introduction}

In recent years, Large Language Models (LLMs) have demonstrated strong capabilities in reasoning, knowledge-intensive tasks, and complex decision-making~\cite{shen-etal-2024-small}, leading to growing adoption in financial applications such as compliance analysis, report summarization, sentiment monitoring, and trading signal extraction~\cite{Europarl2025_AI_FinancialSector}. However, financial services operate under strict privacy regulations, including GDPR and PCI DSS, which restrict the use of third-party cloud APIs for sensitive data. This creates a deployment dilemma: high-performing closed-source models cannot be freely used, while private large-scale infrastructure is prohibitively expensive for most institutions~\cite{wang-etal-2025-unveiling-privacy}.

Although major financial institutions can build private LLM clusters, smaller organizations—such as regional banks, boutique funds, and individual investors—lack the resources to deploy large models locally. In practice, they are constrained to small language models (SLMs) with fewer than 10 billion parameters. While efficient, these models often struggle with multi-step numerical reasoning, domain-specific knowledge, and financial accuracy. Traditional improvement strategies such as large-scale pre-training or task-specific fine-tuning remain costly and impractical in resource-constrained settings~\cite{haque2025tinyllmevaluationoptimizationsmall}.

Recent advances in agent-based systems suggest that tool use and structured collaboration may compensate for model limitations~\cite{shen-etal-2024-small}. However, it remains unclear whether such paradigms effectively enhance small models under strict hardware and privacy constraints. Moreover, existing financial benchmarks primarily evaluate task accuracy, overlooking deployment-oriented factors such as energy consumption, latency, and system robustness.

In this work, we shift the focus from model scaling to system design. We evaluate 27 open-source SLMs across three paradigms: base prompting, tool-augmented single-agent systems, and collaborative multi-agent systems. Experiments span 20 financial datasets across 8 task categories, measuring both effectiveness and resource efficiency. We make the following contributions:
\begin{itemize}[noitemsep, topsep=0pt,leftmargin=*]
\item We present the first large-scale empirical study of $\leq 10$B open-source SLMs in financial settings, systematically comparing base, single-agent, and multi-agent paradigms across 27 models.
\item We provide the first quantitative analysis of the trade-offs introduced by agent-based designs, showing that while tool augmentation improves effectiveness, multi-agent collaboration incurs significant coordination overhead and instability.
\item We provide a practical guide for choosing the best system design based on the available model and specific financial tasks.
\end{itemize}

\section{Related Work and Positioning}
\label{related}

Existing evaluations of large language models focus primarily on reasoning ability and output correctness. Representative benchmarks such as the Open LLM Leaderboard and HELM~\cite{chiang2023open,liang2022holistic} evaluate single models across diverse tasks but assume stable environments with sufficient computational resources. They emphasize final accuracy while largely overlooking deployment factors such as runtime stability, energy cost, and latency. Consequently, these benchmarks provide limited insight into practical usability under resource-constrained or long-running settings.

In finance, LLMs have been widely applied to text analysis and decision support, including FinGPT and BloombergGPT~\cite{yang2023fingpt,wu2023bloomberg}. These approaches typically rely on domain-specific fine-tuning or retrieval augmentation to improve performance. However, most depend on cloud APIs or centralized infrastructure, limiting applicability in localized or privacy-constrained environments.

Agent-based methods offer an alternative path. The Reasoning and Acting (ReAct) framework demonstrates that models can interleave reasoning with tool use to handle complex tasks~\cite{yao2023react}, inspiring multi-agent systems such as AutoGen, LangChain, and CrewAI~\cite{wu2024autogen}. Yet existing studies mainly highlight task completion capability rather than runtime stability or efficiency, and typically rely on large models. Systematic analysis of agent behavior under small-model and resource-constrained settings remains limited.

Recent work has examined the efficiency and optimization of small language models, including latency and system-level improvements~\cite{phametal2025slm,dettmers2023efficient,dao2022flashattention}. However, these studies rarely connect deployment analysis with agent paradigms. As a result, we lack a clear understanding of how agent systems behave under realistic operational constraints.

This work addresses this gap by evaluating whether architectural design can compensate for limited model scale under privacy and hardware constraints, using the financial domain as a practical testbed.

\section{Methodology and Experimental Design}
Our study is designed to systematically evaluate the performance of three distinct paradigms: 1)Base SLMs; 2)Single-Agent Systems (SAS); 3) Multi-Agent Systems (MAS) across a diverse set of financial tasks. This section details our experimental setup, including the selected models, task formulations, and the architectural designs of the single-agent and multi-agent systems.

\subsection{Architecture Designs of Three Paradigms}
\begin{figure*}[!ht]
\includegraphics[width=\textwidth]{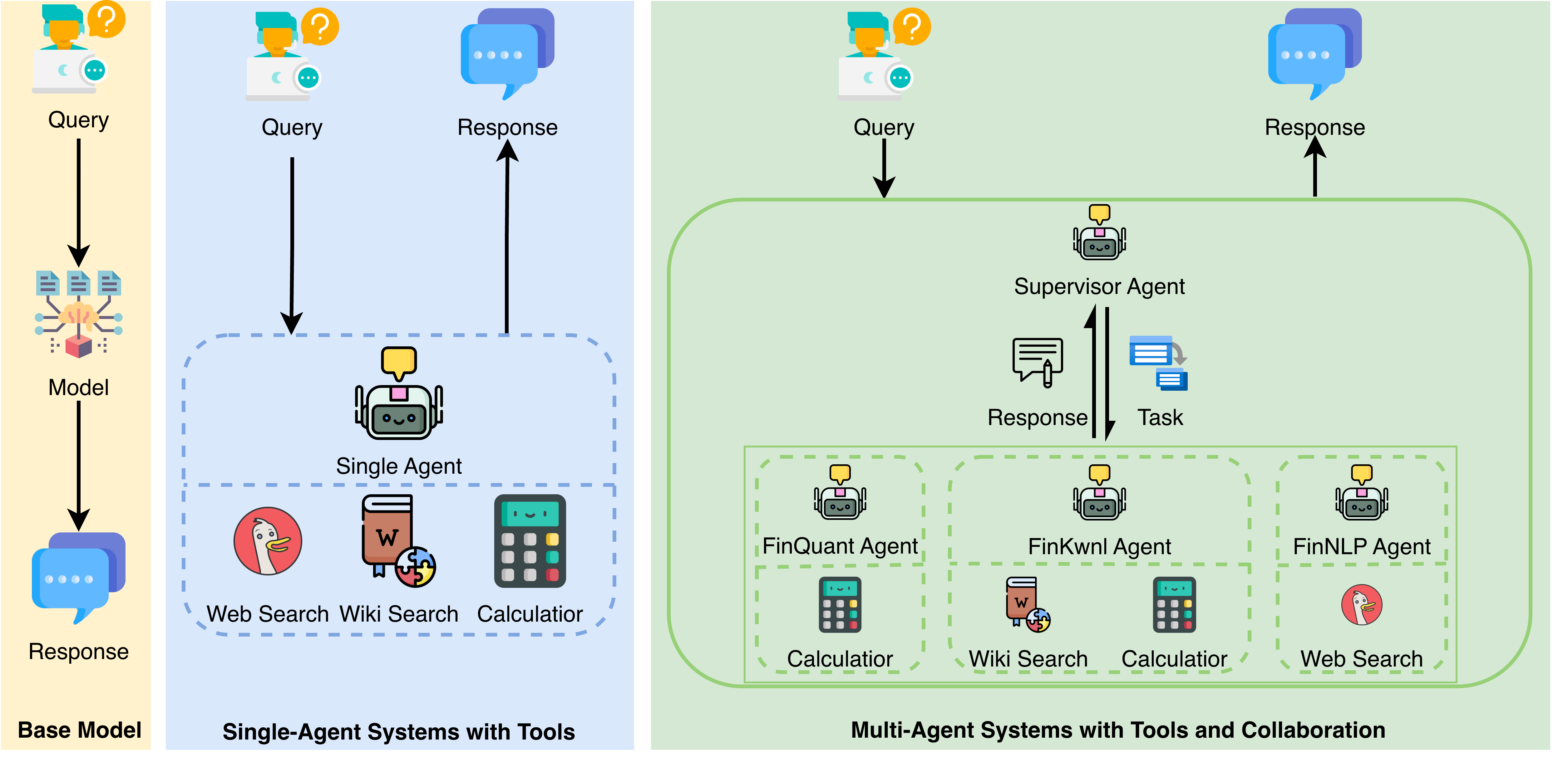}
  \caption{Architecture of three paradigms. The dashed rounded rectangle encloses the agent and the tools it can invoke. The supervisor agent, depicted within a solid rectangle, oversees and invokes the sub-agents contained in the inner solid rounded rectangle.}
  \label{fig:sys}
  \vspace{-15pt}
\end{figure*}

In this study, we consider three progressively complex paradigms, ranging from direct model use to collaborative agent systems. Fig.~\ref{fig:sys} shows their differences in control flow, tool access, and coordination structure.

All agent systems are implemented using the ReAct framework~\cite{yao2022react} (see Appendix~\ref{fig:react}), which alternates between reasoning and tool use. This structure supports step-by-step processing, which is useful for financial tasks. Other frameworks such as Plan-and-Execute~\cite{wang2023plan} and Reflexion~\cite{shinn2023reflexion} are not explored in this study.

\subsubsection{Base SLM}
Base SLM represents direct deployment without architectural changes. Each model receives the task input and generates the output directly. This setting reflects raw model capability and serves as a reference for performance, stability, and resource usage.

\subsubsection{Single-Agent System}




SAS simulates the workflow of an all-round analyst and follows a think–act–observe cycle. After receiving a task, the agent decides whether to use external tools (calculator, wiki search, web search) or answer directly. If tools are used, the agent processes the returned results and may repeat the cycle until producing a final answer. This setup represents a general-purpose financial analyst with tool support.

\subsubsection{Multi-Agent System}



MAS consists of one supervisor and three specialized agents: FinancialKnowledgeAgent, FinancialNLPAgent, and FinancialQuantAgent. All agents follow the ReAct pattern.

The supervisor routes tasks to the most suitable expert agent but does not use tools.
Each expert has limited tool access aligned with its role:
\begin{itemize} [noitemsep, topsep=0pt,leftmargin=*]
    \item FinancialNLPAgent: web search
    \item FinancialKnowledgeAgent: calculator and wiki search
    \item FinancialQuantAgent: calculator only
\end{itemize}
This design introduces role specialization and coordination among agents.

\subsection{Models, Datasets and Tasks}
\subsubsection{Model Selection}
We evaluate 27 open-source language models with fewer than 10 billion parameters. These models~\ref{tab:models} span several widely used families, including Qwen, Gemma, LLaMA, Phi, DeepSeek, Mistral, and Solar, reflecting a diverse range of architectural and training choices commonly considered for local deployment.

\begin{table}[!htpb]
\centering
\setlength{\abovecaptionskip}{2pt}
\begin{tabular}{p{1.3cm}p{3.5cm}l}
    \hline
    \textbf{Model} & \textbf{Version} & \textbf{Range} \\
\hline
Qwen   & 2.5-0.5B, 2.5-1.5B, 2.5-3B, 2.5-7B, 3-8B & 0.5B-8B    \\ 
Llama  & 3.2-1B, 3.2-3B, 2-7B, 3.1-8B, 3-8B       & 1B-8B      \\ 
Gemma  & 3-270M, 3-1B, 2-2B, 3-4B, 2-9B& 0.27B-9B   \\
Phi    & 3.5-mini, 4-mini, 3-small   & 3.8B-7B    \\ 
DeepSeek-R1  & Distill-Qwen-1.5B/7B, Distill-Llama-8B   & 1.5B-8B   \\
Mistral& 7B-v0.3, Ministral-8B   & 7B-8B      \\ 
SOLAR  & 10.7B-v1.0 & 10.7B     \\ 
\hline
\end{tabular}
\caption{Model family, version, and parameter range of evaluated models.}
\label{tab:models}
\end{table}

\subsubsection{Datasets, Tasks and Evaluation Metrics}

Experiments are conducted on 20 publicly available financial datasets covering eight task categories: sentiment analysis, text classification, named entity recognition, question answering, stock movement prediction, credit scoring, summarization, and bankruptcy prediction. These tasks collectively reflect document-level reasoning, numerical understanding, and decision-making within financial contexts.

All datasets provide ground-truth labels for automatic evaluation. A complete dataset summary is provided in Appendix~\ref{tab:dataset}. To ensure a balanced comparison for those tasks, we sample 50 instances from each dataset.

\subsection{Evaluation Metrics for Deployment Reality}

To evaluate agent paradigms under realistic deployment constraints, we adopt a set of metrics that jointly capture effectiveness, efficiency, and robustness. 
Standard task-specific metrics are used where appropriate, but they are not redefined here due to space constraints. Below, we describe the metrics defined for this study.

\paragraph{Completion Rate.} Completion Rate measures the robustness of a system in practical deployment settings.
We define it as the proportion of samples for which the system returns a valid response without runtime errors, timeouts, or malformed outputs:
\[
\text{Completion Rate} = \frac{\# \text{ of successful responses}}{\# \text{ of total samples}}.
\]
This metric reflects the reliability of a paradigm beyond its nominal task performance.

\paragraph{Average Latency.} Average Latency is defined as the mean end-to-end inference time required to process a single input sample, including all intermediate reasoning and agent interactions.
This metric captures user-perceived responsiveness and is critical for latency-sensitive applications.

\paragraph{Normalized Response Quality (NRQ)} To aggregate response quality across heterogeneous tasks and evaluation metrics, we define NRQ as a relative, architecture-level measure (see Appendix~\ref{app:nrq}). 
NRQ captures the overall effectiveness of an architecture while remaining invariant to the scale and type of task-specific evaluation metrics.
For each dataset, response quality is expressed as the normalized improvement over the Base SLM, with the direction adjusted according to whether the underlying metric is higher-is-better or lower-is-better.

\paragraph{Composite Effectiveness Score.} Due to the differences in evaluation metrics among datasets, to enable cross-dataset comparison, we compute a standardized composite score $Z_c$ for each model-architecture combination.

\[
Z_c = \frac{1}{N}\sum_{i=1}^{N} \frac{X_i - \mu_i}{\sigma_i}
\]
where $x$ denotes the raw metric value, and $\mu$ and $\sigma$ denote the mean and standard deviation computed across datasets for the same model. The composite effectiveness Z-score is then obtained by averaging the Z-scores across all available dataset-specific metrics. A higher $Z_c$ indicates better overall effectiveness relative to other configurations.

\paragraph{Leading Advantage.} To quantify the decisiveness of performance differences, we define leading advantage as the relative gap between the best-performing and second-best-performing architectures :
\[
\alpha = \frac{s_{\text{best}} - s_{\text{second}}}{|s_{\text{second}}| + \epsilon} \times 100\%, \quad \epsilon=10^{-8}
\]
where $\epsilon$ is a small constant used to avoid division by zero.
This metric highlights whether observed improvements represent marginal gains or substantial performance advantages.

\subsection{Implementation}
All experiments were conducted in a unified environment to ensure comparability. We implement the experiments on NVIDIA H100 80GB GPU, CUDA 12.1, to ensure finish the experiments quickly. Regarding the inference parameters setting, temperature was set to 0 to reduce randomness and enhance reproducibility, with top\_p=0.9. The maximum interaction turns of agentic architectures were capped at 5 considering the time efficiency. We used vLLM for high-throughput, memory-efficient inference on HPC.

\section{Results and Analysis}
\subsection{Overall System Trade-off}

Table~\ref{tab:overview} summarizes architecture-level performance across effectiveness, efficiency, and stability, revealing a clear trade-off as system complexity increases.

In terms of effectiveness, SAS achieves the highest NRQ (4.85), showing clear improvement over the Base SLM. MAS provides only limited additional gains (NRQ = 0.36). However, reliability declines with complexity. The Base SLM maintains a near-perfect completion rate (99.67\%), which drops to 79.92\% for SAS and 72.01\% for MAS. This suggests that more complex architectures introduce instability during execution.

Regarding efficiency, energy per token decreases from 1.83 (Base) to 1.03 (SAS) and 0.53 (MAS), a 71\% reduction from Base to MAS. Tokens per second also increase (190 → 345 → 642). However, average latency nearly doubles for SAS and MAS (about 23 seconds) compared to the Base SLM (12.39 seconds). Token usage rises sharply from 2,356 tokens per sample (Base) to 8,040 (SAS) and 14,618 (MAS), reflecting coordination and multi-step reasoning overhead.

\begin{table*}[!t]
\centering
\setlength{\abovecaptionskip}{2pt}
\begin{tabular}{cccccccc}
\hline
\textbf{Architecture} & \textbf{Comp. Rate$\uparrow$} & \textbf{NRQ$\uparrow$} & \textbf{Energy/T(mJ)$\downarrow$} & \textbf{Avg. Latency(s)$\downarrow$} & \textbf{T/s$\uparrow$} & \textbf{T/Sample$\downarrow$} \\
\hline
Base& \textbf{99.67\%} & 0.00          & 1.83          & \textbf{12.39} & 190.08 & \textbf{2356.06} \\
SAS & 79.92\%          & \textbf{4.85} & 1.02          & 23.27          & 345.42 & 8039.75 \\
MAS & 72.01\%          & 0.36          & \textbf{0.52} & 22.76          & \textbf{642.34} & 14618.22 \\
\hline  
\end{tabular}
\caption{Overall architecture-level comparison across effectiveness, efficiency, and system stability. Arrows indicate optimization direction ($\uparrow$ higher is better; $\downarrow$ lower is better). T denotes tokens. s denotes second.}
\label{tab:overview}
\vspace{-15pt}
\end{table*}

\subsection{Efficiency–Effectiveness Trade-off}
\label{subsec:tradeoff}

\begin{figure}[!ht]
\setlength{\abovecaptionskip}{2pt}
  \includegraphics[width=\columnwidth]{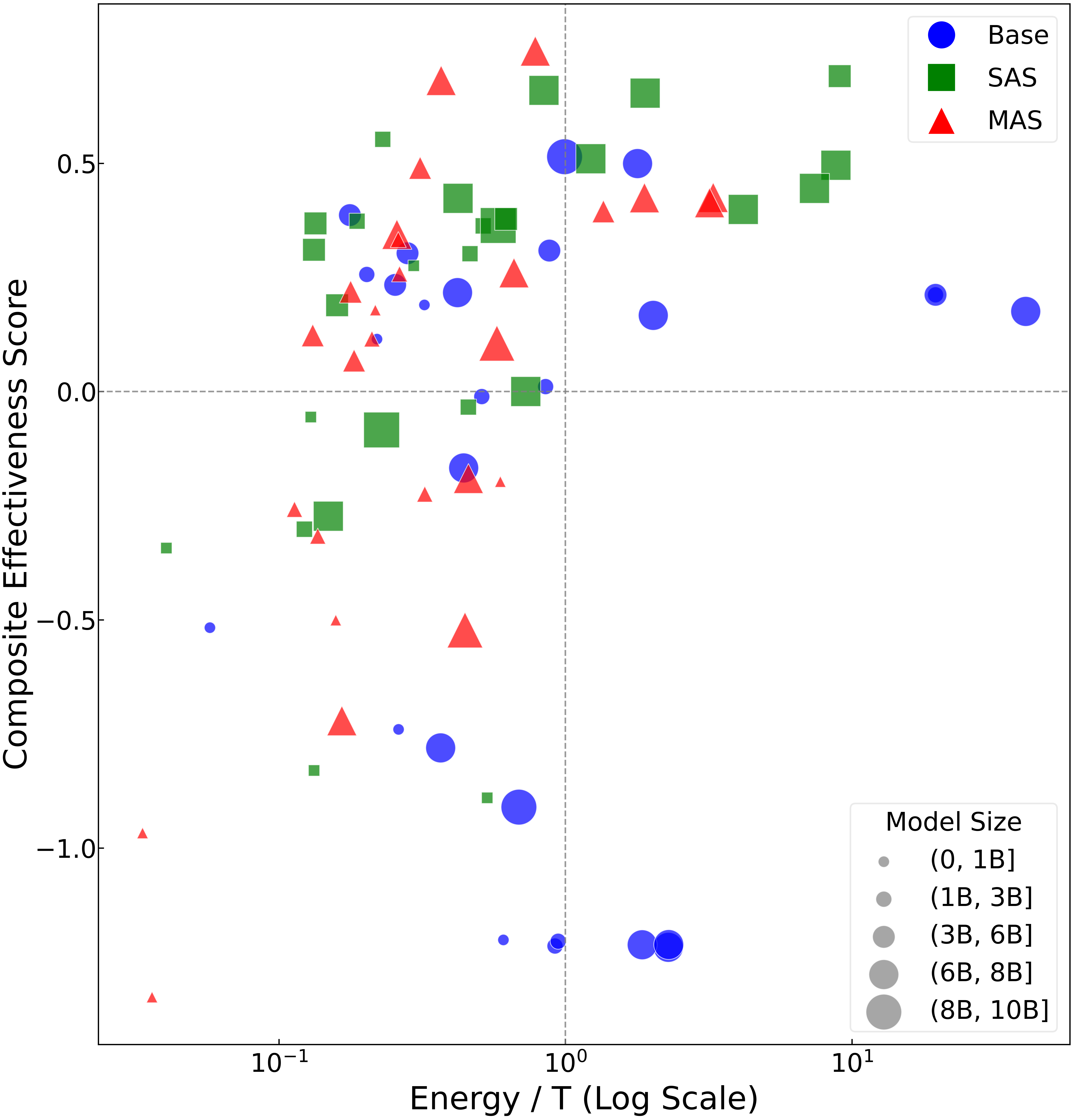}
  \caption{Efficiency-Effectiveness trade-off across model-architecture settings. Each point represents one model under one architecture. Different marker shapes indicate different architectures, and marker size reflects model parameter size.}
  \label{fig:ee}
\end{figure}

To evaluate practical utility, we analyze the trade-off between computational efficiency ($\text{Energy}/T$, log scale) and effectiveness (composite effectiveness score). As illustrated in Fig.~\ref{fig:ee}, our empirical results reveal a clear divergence in the Pareto frontier across the Base, SAS, and MAS. 


SAS (green squares) consistently occupies the upper even the upper-left quadrant of the trade-off space, achieving strong effectiveness (mostly $z>0.4$) at relatively low energy levels. This indicates that single-agent design improves performance without large coordination cost.


In contrast, Base SLMs (blue circles) show an efficiency trap at larger scales. While smaller models perform moderately, $8\text{B}$--$10\text{B}$ variants move toward the lower-right region, combining high energy use with lower effectiveness ($z<-0.5$). This suggests that scaling parameter volume in isolation may yield diminishing returns. MAS (red triangles) shows higher variance. It can reach high effectiveness but often requires more energy due to coordination overhead, and smaller MAS setups sometimes perform worse than SAS.


\subsection{Task–Architecture Adaptation}
\begin{figure*}[!htbp]
\setlength{\abovecaptionskip}{2pt}
  \includegraphics[width=\textwidth]{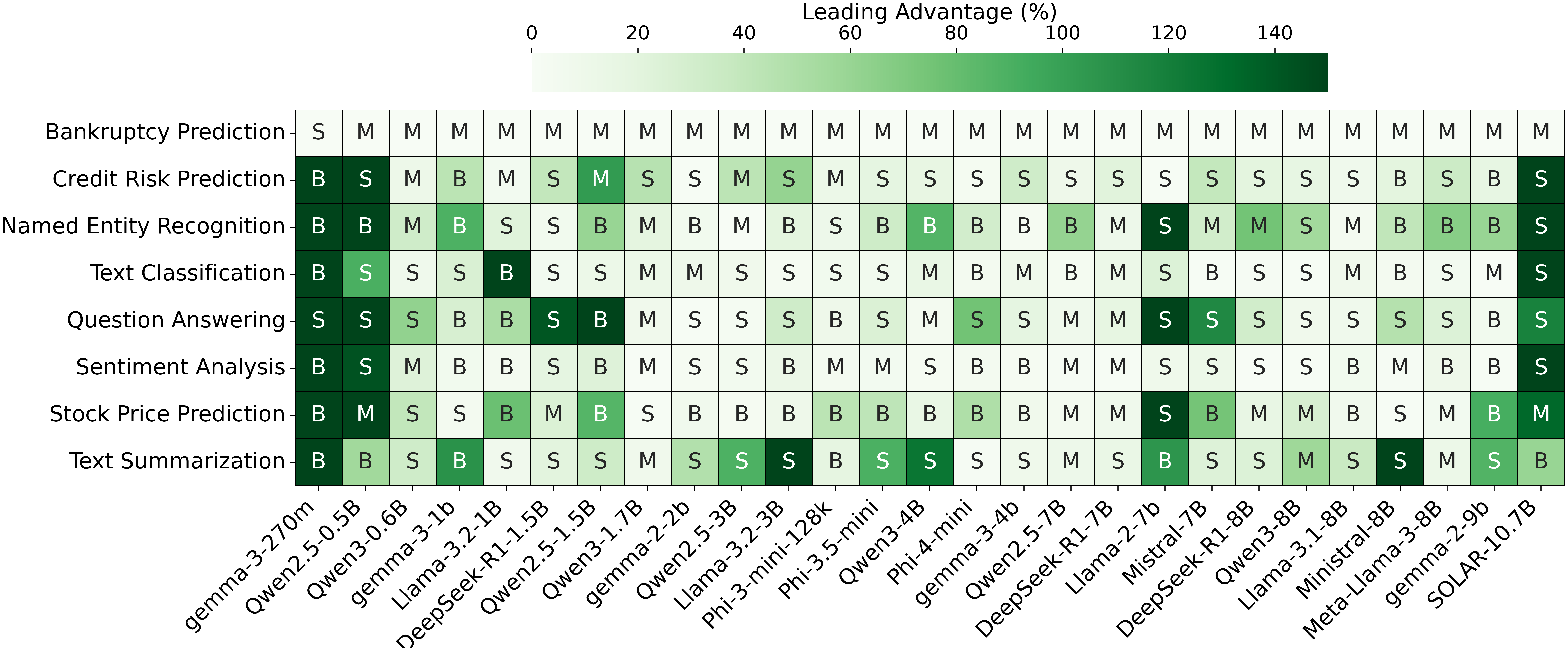}
  \caption{Heatmap of task-architecture adaption. B denotes Base SLM; S denotes SAS; M denotes MAS. The letter in each cell means that setting performs best. The background color of each cell shows how large the advantage is. Darker colors indicate a larger performance gap between the best and second-best options.}
  \label{fig:pattern}
    \vspace{-15pt}
\end{figure*}

Beyond global metrics, we analyze how Base (B), SAS (S), and MAS (M) perform across different tasks (Fig.~\ref{fig:pattern}). Results show that no single paradigm dominates across all settings; the best choice depends on both the model family and task type.


MAS performs best on bankruptcy prediction across most model scales, though often with a small margin. This suggests that multi-agent coordination can benefit high-risk financial tasks, despite its higher cost. In contrast, SAS works well for reasoning and generation tasks such as question answering, summarization, and credit risk prediction. For example, in Qwen2.5-0.5B and Llama-2-7b, SAS achieves over ($>150\%$) improvement compared to other setups, showing that single-agent design can strengthen mid-sized models.


Interestingly, Base SLMs remain competitive for classification and token-level tasks, including named entity recognition, sentiment analysis, and stock prediction, especially in the Gemma and Phi families. For gemma-3-270m, the Base SLM outperforms agent paradigms by a large margin on most tasks. This is because the agent paradigms of this model fail to run on most tasks.

\subsection{Failure Mode Analysis}

\begin{figure}[!htpb]
\setlength{\abovecaptionskip}{2pt}
  \includegraphics[width=\columnwidth]{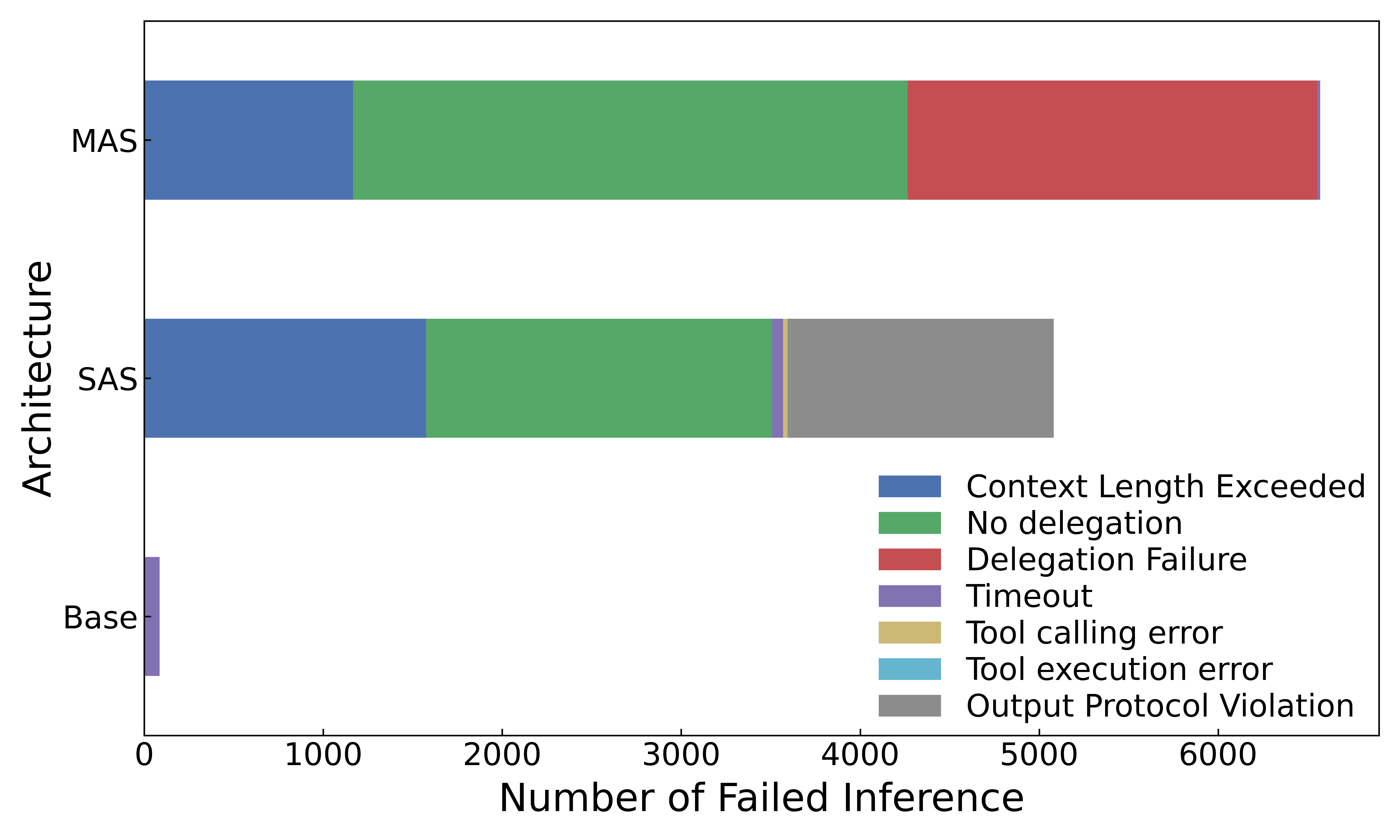}
  \caption{Distribution of failure modes for Base, SAS and MAS}
  \label{fig:failure}
\end{figure}

To compare direct inference and agentic orchestration, we analyze failure modes in Base, SAS, and MAS. Fig.~\ref{fig:failure} shows that agentic systems (SAS and MAS) have more errors than the Base SLM. This is because the Base SLM either produces an output or times out, while SAS and MAS use multi-turn reasoning and tools, which increases system-level failures.

ReAct in SAS and MAS shifts failures toward structural and budget issues. In SAS, \textit{Context Length Exceeded} and \textit{No Delegation dominate}, as long prompts and iterative reasoning exhaust the context window. In MAS, inter-agent dynamics cause \textit{Delegation Failure} and \textit{No Delegation}. While MAS avoids \textit{Timeout} and\textit{ Output Protocol Violations}, agents can get stuck in failed hand-offs or fail to start delegation, offsetting the benefits of the multi-agent setup.





\section{Discussion and Practical Implications}
\subsection{Discussion}

Our results challenge the prevailing "scaling law" hypothesis~\cite{kaplan2020scaling}, which posits that performance is a monotonic function of parameter count and computation. The observed "U-turn" in Base SLMs (Fig.~\ref{fig:ee}) reveals an optimization bottleneck where increased model capacity without an agent paradigm yields diminishing, and eventually negative, returns. However, the transition to agentic architectures introduces a fundamental coordination tax~\cite{zhang2024cutcrapeconomicalcommunication,rizvimartel2025benefitslimitationscommunicationmultiagent}. While Single-Agent Systems offer a Pareto-optimal balance by concentrating reasoning capabilities with moderate energy costs, Multi-Agent Systems suffer from the energy expenditure of inter-agent communication, and iterative ReAct prompting frequently fails to translate into proportional gains in correctness. 

This decoupling of efficiency and effectiveness highlights a critical form of architectural fragility. By wrapping models in complex agent paradigms, we trade the unpredictability of raw model outputs for the complexity of managing agent states. The high incidence of \textit{Delegation Failure} and \textit{Context Length Exceeded} in our failure analysis suggests that the bottleneck for modern AI is no longer raw generative capability but rather context management and instruction adherence. Consequently, increasing agent counts often introduces new vectors for systemic failure, such as infinite delegation loops, rather than enhancing collective intelligence.



\subsection{Implications for Industry Deployment}
From a practical view, these findings show that we should stop thinking that "bigger is always better." Instead, we should choose the design that fits the specific task.

First, we need to choose the paradigm based on what it does best. SAS is the best choice for complex creative tasks because it works very well and saves energy. While the high variance of MAS suggests it should be restricted to high-entropy domains, such as financial forecasting, where the statistical redundancy of multiple perspectives outweighs the coordination costs. For simpler extraction tasks like named entity recognition, the Base remains superior, as agentic overhead tends to dilute precise token-level mappings.

Second, these systems break easily, so we need to build them carefully. Builders should make backup plans. If the system gets stuck or runs out of space, it should switch back to the basic model to ensure it still provides an answer.

\subsection{Limitations and Future Work}
This study covers a fixed set of small language models, tasks, and agent designs. Other settings may lead to different results. The agent systems use simple coordination strategies, and stronger control mechanisms may improve stability. Adaptive agents that change behavior over time are not included. Future work could study dynamic control, better delegation strategies, and system behavior under real user traffic.


\bibliography{main}

@report{Europarl2025_AI_FinancialSector,
  author       = {European Parliament, Committee on Economic and Monetary Affairs and Rapporteur Arba Kokalari},
  title        = {Report on the impact of artificial intelligence on the financial sector (A10-0225/2025)},
  institution  = {European Parliament},
  year         = {2025},
  month        = {November},
  day          = {11},
  url          = {https://www.europarl.europa.eu/doceo/document/A-10-2025-0225_EN.html},
  note         = {Adopted 5.11.2025 by the Committee on Economic and Monetary Affairs}
}

@article{kaplan2020scaling,
  title={Scaling Laws for Neural Language Models},
  author={Kaplan, Jared and McCandlish, Sam and Henighan, Tom and Brown, Tom B and Chess, Benjamin and Child, Rewon and Gray, Scott and Radford, Alec and Wu, Jeffrey and Amodei, Dario},
  journal={arXiv preprint arXiv:2001.08361},
  year={2020}
}

@article{shinn2023reflexion,
  title={Reflexion: Language agents with verbal reinforcement learning, 2023},
  author={Shinn, Noah and Cassano, Federico and Labash, Beck and Gopinath, Ashwin and Narasimhan, Karthik and Yao, Shunyu},
  journal={URL https://arxiv. org/abs/2303.11366},
  volume={1},
  year={2023}
}

@inproceedings{yao2022react,
  title={React: Synergizing reasoning and acting in language models},
  author={Yao, Shunyu and Zhao, Jeffrey and Yu, Dian and Du, Nan and Shafran, Izhak and Narasimhan, Karthik R and Cao, Yuan},
  booktitle={The eleventh international conference on learning representations},
  year={2022}
}

@article{wang2023plan,
  title={Plan-and-Solve Prompting: Improving Zero-Shot Chain-of-Thought Reasoning by Large Language Models},
  author={Wang, Lei and Xu, Wanyu and Lan, Yihuai and Hu, Zhiqiang and Lan, Yunshi and Lee, Roy Ka-Wei and Lim, Ee-Peng},
  journal={arXiv preprint arXiv:2305.04091},
  year={2023}
}

@inproceedings{liang2022holistic,
  title     = {Holistic Evaluation of Language Models},
  author    = {Liang, Percy and Bommasani, Rishi and Lee, Tony and others},
  booktitle = {Advances in Neural Information Processing Systems (NeurIPS)},
  year      = {2022}
}

@inproceedings{chiang2023open,
  title     = {Open LLM Leaderboard},
  author    = {Chiang, Wei-Lin and others},
  booktitle = {NeurIPS Datasets and Benchmarks Track},
  year      = {2023}
}

@inproceedings{yang2023fingpt,
  title     = {FinGPT: Open-Source Financial Large Language Models},
  author    = {Yang, Shuo and Liu, Yilun and others},
  booktitle = {Proceedings of the 2023 Conference on Empirical Methods in Natural Language Processing (EMNLP), Industry Track},
  year      = {2023}
}

@inproceedings{wu2023bloomberg,
  title     = {BloombergGPT: A Large Language Model for Finance},
  author    = {Wu, Shijie and Irsoy, Ozan and others},
  booktitle = {Proceedings of the 61st Annual Meeting of the Association for Computational Linguistics (ACL)},
  year      = {2023}
}

@inproceedings{yao2023react,
  title     = {ReAct: Synergizing Reasoning and Acting in Language Models},
  author    = {Yao, Shunyu and Zhao, Jeffrey and others},
  booktitle = {International Conference on Learning Representations (ICLR)},
  year      = {2023}
}

@inproceedings{wu2024autogen,
  title     = {AutoGen: Enabling Next-Gen LLM Applications via Multi-Agent Conversation},
  author    = {Wu, Qingyun and others},
  booktitle = {International Conference on Learning Representations (ICLR)},
  year      = {2024}
}

@inproceedings{dettmers2023efficient,
  title     = {Efficient Fine-Tuning of Small Language Models},
  author    = {Dettmers, Tim and others},
  booktitle = {International Conference on Machine Learning (ICML)},
  year      = {2023}
}

@inproceedings{dao2022flashattention,
  title     = {FlashAttention: Fast and Memory-Efficient Exact Attention},
  author    = {Dao, Tri and others},
  booktitle = {Advances in Neural Information Processing Systems (NeurIPS)},
  year      = {2022}
}

@inproceedings{phametal2025slm,
    title = "{SLM}-Bench: A Comprehensive Benchmark of Small Language Models on Environmental Impacts",
    author = "Pham, Nghiem Thanh  and
      Kieu, Tung  and
      Nguyen, Duc Manh  and
      Xuan, Son Ha  and
      Duong-Trung, Nghia  and
      Le-Phuoc, Danh",
    editor = "Christodoulopoulos, Christos  and
      Chakraborty, Tanmoy  and
      Rose, Carolyn  and
      Peng, Violet",
    booktitle = "Findings of the Association for Computational Linguistics: EMNLP 2025",
    month = nov,
    year = "2025",
    address = "Suzhou, China",
    publisher = "Association for Computational Linguistics",
    url = "https://aclanthology.org/2025.findings-emnlp.1165/",
    doi = "10.18653/v1/2025.findings-emnlp.1165",
    pages = "21369--21392",
    ISBN = "979-8-89176-335-7",
}

@inproceedings{shen-etal-2024-small,
    title = "Small {LLM}s Are Weak Tool Learners: A Multi-{LLM} Agent",
    author = "Shen, Weizhou  and
      Li, Chenliang  and
      Chen, Hongzhan  and
      Yan, Ming  and
      Quan, Xiaojun  and
      Chen, Hehong  and
      Zhang, Ji  and
      Huang, Fei",
    editor = "Al-Onaizan, Yaser  and
      Bansal, Mohit  and
      Chen, Yun-Nung",
    booktitle = "Proceedings of the 2024 Conference on Empirical Methods in Natural Language Processing",
    month = nov,
    year = "2024",
    address = "Miami, Florida, USA",
    publisher = "Association for Computational Linguistics",
    url = "https://aclanthology.org/2024.emnlp-main.929/",
    doi = "10.18653/v1/2024.emnlp-main.929",
    pages = "16658--16680"
}

@inproceedings{wang-etal-2025-unveiling-privacy,
    title = "Unveiling Privacy Risks in {LLM} Agent Memory",
    author = "Wang, Bo  and
      He, Weiyi  and
      Zeng, Shenglai  and
      Xiang, Zhen  and
      Xing, Yue  and
      Tang, Jiliang  and
      He, Pengfei",
    editor = "Che, Wanxiang  and
      Nabende, Joyce  and
      Shutova, Ekaterina  and
      Pilehvar, Mohammad Taher",
    booktitle = "Proceedings of the 63rd Annual Meeting of the Association for Computational Linguistics (Volume 1: Long Papers)",
    month = jul,
    year = "2025",
    address = "Vienna, Austria",
    publisher = "Association for Computational Linguistics",
    url = "https://aclanthology.org/2025.acl-long.1227/",
    doi = "10.18653/v1/2025.acl-long.1227",
    pages = "25241--25260",
    ISBN = "979-8-89176-251-0"
}

@misc{haque2025tinyllmevaluationoptimizationsmall,
      title={TinyLLM: Evaluation and Optimization of Small Language Models for Agentic Tasks on Edge Devices}, 
      author={Mohd Ariful Haque and Fahad Rahman and Kishor Datta Gupta and Khalil Shujaee and Roy George},
      year={2025},
      eprint={2511.22138},
      archivePrefix={arXiv},
      primaryClass={cs.LG},
      url={https://arxiv.org/abs/2511.22138}, 
}

@misc{zhang2024cutcrapeconomicalcommunication,
      title={Cut the Crap: An Economical Communication Pipeline for LLM-based Multi-Agent Systems}, 
      author={Guibin Zhang and Yanwei Yue and Zhixun Li and Sukwon Yun and Guancheng Wan and Kun Wang and Dawei Cheng and Jeffrey Xu Yu and Tianlong Chen},
      year={2024},
      eprint={2410.02506},
      archivePrefix={arXiv},
      primaryClass={cs.MA},
      url={https://arxiv.org/abs/2410.02506}, 
}

@misc{rizvimartel2025benefitslimitationscommunicationmultiagent,
      title={Benefits and Limitations of Communication in Multi-Agent Reasoning}, 
      author={Michael Rizvi-Martel and Satwik Bhattamishra and Neil Rathi and Guillaume Rabusseau and Michael Hahn},
      year={2025},
      eprint={2510.13903},
      archivePrefix={arXiv},
      primaryClass={cs.MA},
      url={https://arxiv.org/abs/2510.13903}, 
}

@misc{xie2023pixiu,
      title={PIXIU: A Large Language Model, Instruction Data and Evaluation Benchmark for Finance}, 
      author={Qianqian Xie and Weiguang Han and Xiao Zhang and Yanzhao Lai and Min Peng and Alejandro Lopez-Lira and Jimin Huang},
      year={2023},
      eprint={2306.05443},
      archivePrefix={arXiv},
      primaryClass={cs.CL}
}

@misc{xie2024FinBen,
      title={The FinBen: An Holistic Financial Benchmark for Large Language Models}, 
      author={Qianqian Xie and Weiguang Han and Zhengyu Chen and Ruoyu Xiang and Xiao Zhang and Yueru He and Mengxi Xiao and Dong Li and Yongfu Dai and Duanyu Feng and Yijing Xu and Haoqiang Kang and Ziyan Kuang and Chenhan Yuan and Kailai Yang and Zheheng Luo and Tianlin Zhang and Zhiwei Liu and Guojun Xiong and Zhiyang Deng and Yuechen Jiang and Zhiyuan Yao and Haohang Li and Yangyang Yu and Gang Hu and Jiajia Huang and Xiao-Yang Liu and Alejandro Lopez-Lira and Benyou Wang and Yanzhao Lai and Hao Wang and Min Peng and Sophia Ananiadou and Jimin Huang},
      year={2024},
      eprint={2402.12659},
      archivePrefix={arXiv},
      primaryClass={cs.CL}
}

\onecolumn
\newpage
\twocolumn

\appendix
\section{Details of Reasoning and Acting Implementation}
\label{fig:react}
\begin{figure}[!htpb]
\includegraphics[width=\columnwidth]{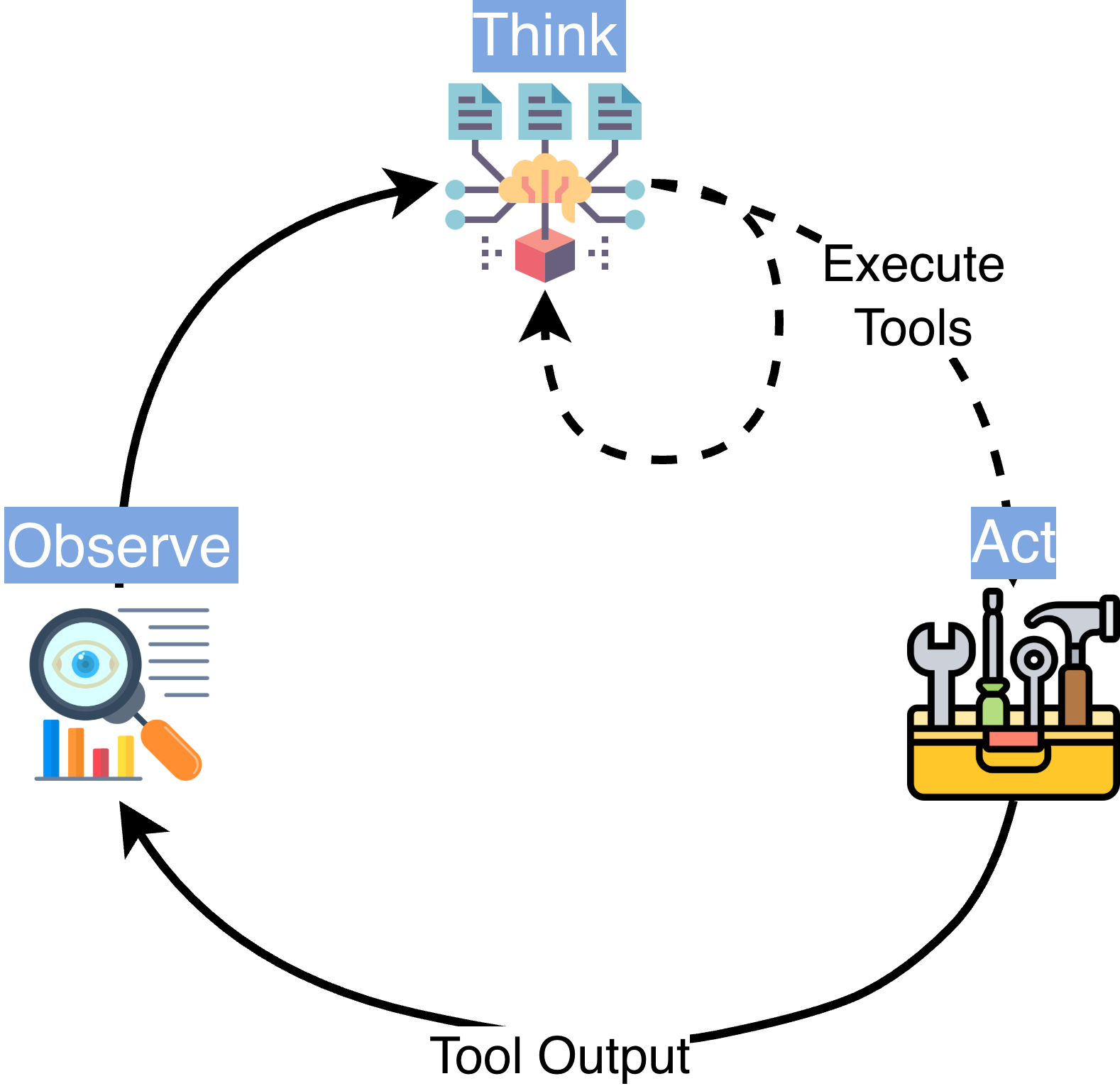}
  \caption{Structure of Reasoning and act}
  \label{react}
\end{figure}

As illustrated by the think-act-observe cycle~\ref{react}, upon receiving a task, the agent first enters the thinking phase, autonomously determining whether to invoke tools (calculator, wiki search, or web search) and their invocation sequence. The agent may provide answers directly without tool assistance. Should it identify a tool requirement, it proactively transitions to the action phase to invoke external tools, then proceeds to the observation phase: ingesting tool outputs, validating them against intermediate reasoning, and updating its internal state.
This cycle (think-act-observe) may iterate multiple times until the model accumulates sufficient empirical information to synthesize the final answer. 


\section{Summary of Dataset}
\label{tab:dataset}
Our evaluation spans eight representative financial NLP tasks, selected to cover a broad range of linguistic, analytical, and decision-making challenges encountered in real-world financial applications. These tasks include sentiment analysis, text classification, named entity recognition, question answering, stock movement prediction, credit scoring, and two financial summarization tasks. Together, they reflect both document-level and sentence-level reasoning, factual grounding, numerical understanding, and predictive modeling.

To ensure a comprehensive assessment, we evaluate performance across 20 publicly available datasets as shown in Table~\ref{dataset}, each widely used in the financial NLP community. The datasets vary in size—from small expert-annotated corpora to large-scale benchmark collections—and target distinct aspects of financial language understanding, such as investor sentiment (FiQA SA, Financial PhraseBank), numerical table reasoning (FinQA, TATQA), risk classification (German, Australian, LendingClub), and multi-document summarization (ECTSum, EDTSum).

\begin{table*}[!ht]
\begin{tabular}{llcc}
\hline
\multicolumn{1}{l}{\textbf{Tasks}} & \textbf{Dataset}  & \textbf{\# of sample}    & \textbf{Evaluation Metrics}\\ \hline
 & textbf{FiQA SA} & { 235}   & { F1}       \\ \cline{2-4} 
 & { Financial PhraseBank} & { 970}   & { F1}     \\ \cline{2-4} 
 & { TFNS}    & { 2390}  & { F1}     \\ \cline{2-4} 
\multirow{-4}{*}{Sentiment analysis}& { NWGI}    & { 4.05k} & { F1}     \\ \hline
\multicolumn{1}{l}{Text   classification}      & { Headline}& { 20.5k} & { F1}     \\ \hline
\multicolumn{1}{l}{Named entity   recognition} & { NER}     & { 3.5k}  & { F1}     \\ \hline
 & CFA\_QA  & 1032    & {EM Accuracy}  \\ \cline{2-4} 
 & { FinQA}   & { 1147}  & { EM Accuracy}      \\ \cline{2-4} 
 & { ConvFinQA} & { 1490}  & { EM Accuracy}      \\ \cline{2-4} 
\multirow{-4}{*}{Question answering}& { TATQA}   & { 1668}  & {EM Accuracy}   \\ \hline
 & { BigData22}  & { 1470}  & { Accuracy}    \\ \cline{2-4} 
 & { ACL18}   & { 3720}  & { Accuracy}    \\ \cline{2-4} 
\multirow{-3}{*}{Stock movement prediction}      & { CIKM18}  & { 1140}  & { Accuracy}    \\ \hline
 & { German}  & { 200}   & { F1}  \\ \cline{2-4} 
 & { Australian}& { 139}   & { F1}  \\ \cline{2-4} 
\multirow{-3}{*}{Credit Scoring}    & { LendingClub}          & { 2691}  & { F1}  \\ \hline
 & { ECTSum}  & { 495}   & { ROUGE-L} \\ \cline{2-4} 
\multirow{-2}{*}{Summerization}     & { EDTSum}  & { 2000}  & { ROUGE-L} \\ \hline
 & { Polish}  & { 235}   & { AUC} \\ \cline{2-4} 
\multirow{-2}{*}{Bankruptcy prediction}     & { Taiwan}  & { 452}  & { AUC} \\ \hline
\end{tabular}
\caption{Summary of Dataset}
\label{dataset}
\end{table*}

\section{Formal Definition of Normalized Response Quality}
\label{app:nrq}

This section provides the formal definition of the \emph{Normalized Response Quality (NRQ)} metric used in the main paper.
NRQ is designed to aggregate response quality across heterogeneous tasks and datasets with different evaluation metrics while isolating the effect of architectural design choices.

\subsection{Dataset-level Relative Response Quality Gain}

Let $d$ denote a dataset and $a \in \mathcal{A}$ denote an architecture.
We use $s_{a,d}$ to represent the task-specific evaluation score obtained by architecture $a$ on dataset $d$.
Each dataset is associated with a flag indicating whether its evaluation metric is \emph{higher-is-better} (e.g., Accuracy, F1, ROUGE, Exact Match) or \emph{lower-is-better} (e.g., error rate, mean absolute error).

We designate the base architecture as an anchor and compute the relative response quality gain of architecture $a$ on dataset $d$ as:
\[
g_{a,d} =
\begin{cases}
\dfrac{s_{a,d} - s_{\text{Base},d}}
{|s_{\text{Base},d}| + \epsilon},
& \text{if higher-is-better}, \\[10pt]
\dfrac{s_{\text{Base},d} - s_{a,d}}
{|s_{\text{Base},d}| + \epsilon},
& \text{if lower-is-better},
\end{cases}
\]
where $\epsilon = 10^{-8}$ is a small constant used to avoid division by zero.

By construction, $g_{a,d} > 0$ indicates that architecture $a$ improves response quality relative to the base architecture on dataset $d$, while $g_{a,d} < 0$ indicates degradation.

\subsection{Task-level Aggregation}

To prevent tasks with more datasets from disproportionately influencing the aggregate score, we first aggregate relative response quality gains at the task level.
Let $\mathcal{D}_t$ denote the set of datasets associated with task $t$.
The task-level response quality score for architecture $a$ is defined as:
\[
G_{a,t} =
\frac{1}{|\mathcal{D}_t|}
\sum_{d \in \mathcal{D}_t}
g_{a,d}.
\]

This aggregation captures the average architectural effect within a coherent task category, independent of the number of datasets or the scale of their underlying evaluation metrics.

\subsection{Architecture-level Aggregation}

Finally, we compute the overall normalized response quality of architecture $a$ by averaging task-level scores across all tasks.
Let $\mathcal{T}$ denote the set of evaluated tasks.
The architecture-level NRQ score is defined as:
\[
\text{NRQ}(a) =
\frac{1}{|\mathcal{T}|}
\sum_{t \in \mathcal{T}}
G_{a,t}.
\]

This equal-weighted aggregation ensures that all tasks contribute uniformly to the final score, yielding a single architecture-level measure of response quality that is comparable across heterogeneous tasks, datasets, and evaluation metrics.

\section{Evaluation Metrics}
\paragraph{Energy per Token.}
Energy per Token quantifies the computational efficiency of a model by measuring the average energy consumption normalized by the number of generated tokens.
It captures the marginal energy cost of text generation and is particularly relevant for large-scale or resource-constrained deployments.

\paragraph{Tokens per Sample.}
Tokens per Sample measures the average number of tokens generated for each input sample. This metric reflects verbosity and computational load, and serves as a proxy for inference cost in token-based billing or throughput-limited systems.

\section{Prompt Design for Base SLM, Single-Agent and Multi-Agent Systems}

\subsection{Base SLM}
There is no advance prompt engineering regarding the Base SLMs. The prompts for each query are from the datasets in The FinAI community~\cite{xie2023pixiu, xie2024FinBen}.

\subsection{Design Principles for Agent Paradigms}
The prompt design enforces deterministic structured outputs, constrained tool invocation, and explicit reasoning control. The SAS evaluates unified reasoning capacity, while the MAS evaluates decomposed coordination under identical operational constraints.
All prompts are designed under three principles to ensure stable deployment and fair architectural comparison:

\begin{enumerate}[noitemsep, topsep=0pt]
    \item \textbf{Strict structural control.} All responses must follow predefined XML-style tags (e.g., \texttt{<think>}, \texttt{<action>}, \texttt{<observation>}, \texttt{<final\_answer>}) to prevent format drift.
    \item \textbf{Explicit tool governance.} Tool usage is constrained by hard formatting rules to avoid hallucinated tools and malformed JSON calls.
    \item \textbf{Task-grounded reasoning.} The model must first identify the required answer format (e.g., ``yes/no'', ``A/B/C'') before reasoning.
\end{enumerate}

The Single-Agent System (SAS) and Multi-Agent System (MAS) differ only in architectural decomposition, while maintaining identical tools and structural constraints.

\subsection{Single-Agent System (SAS)}

The SAS defines a unified financial intelligence agent specialized in analyzing financial data and answering questions including financial NLP tasks (sentiment, NER, classification, summarization), conceptual finance questions, numerical reasoning, credit scoring and bankruptcy prediction, stock movement prediction.

\subsubsection{Structured Workflow}

The SAS follows a ReAct-style protocol:

\begin{verbatim}
<question>
<think>
<action>
<observation>
<final_answer>
\end{verbatim}

This structure separates reasoning from final output, enforces explicit tool invocation, and ensures that tool results are system-provided rather than hallucinated.

\subsubsection{Tool Governance}

Tool usage rules are explicitly defined:
\begin{itemize}[noitemsep, topsep=0pt]
    \item Mathematical calculations $\rightarrow$ calculator
    \item Concept definitions $\rightarrow$ wikisearcher
    \item Latest information $\rightarrow$ websearcher
    \item Pure reasoning $\rightarrow$ no tool
\end{itemize}

Strict JSON templates prevent nested structures and malformed tool calls, which are common failure modes in tool-augmented LLM systems.

\subsubsection{Minimal SAS Example}

\begin{verbatim}
You are a financial intelligence agent.

Workflow:
1. Read <question>
2. Reason in <think>
3. Call tool in <action> if needed
4. Wait for <observation>
5. Output <final_answer>

Response format:
<question>...</question>
<think>...</think>
<action>...</action>
<observation>...</observation>
<final_answer>...</final_answer>
\end{verbatim}

\subsection{Multi-Agent System (MAS)}

The MAS decomposes reasoning into \textbf{Supervisor}, \textbf{FinancialKnowledgeAgent}, \textbf{FinancialNLPAgent} , and \textbf{FinancialQuantAgent}.

\subsubsection{Supervisor Protocol}

The supervisor never solves tasks directly. It follows a strict two-phase protocol:

\paragraph{Phase 1 (No Observation Present).}
\begin{itemize}[noitemsep, topsep=0pt]
    \item Select exactly one specialist agent.
    \item Delegate using \texttt{<action>}.
    \item Must not output \texttt{<final\_answer>}.
\end{itemize}

\paragraph{Phase 2 (Observation Present).}
\begin{itemize}[noitemsep, topsep=0pt]
    \item Evaluate the agent result.
    \item Output the final answer.
    \item No agent simulation or modification of observation.
\end{itemize}

\textbf{Minimal Supervisor Example:}

\begin{verbatim}
<question>...</question>
<think>Select appropriate agent</think>
<action>{"agent": "financialQuantAgent", 
          "task": "..."}</action>
\end{verbatim}

After receiving observation:

\begin{verbatim}
<think>Evaluate agent result</think>
<final_answer>yes</final_answer>
\end{verbatim}

\subsubsection{Specialist Agent Design}

Each agent has a restricted scope:

\begin{itemize} [noitemsep, topsep=0pt]
    \item \textbf{FinancialKnowledgeAgent:} CFA, financial concepts, numerical reasoning
    \item \textbf{FinancialNLPAgent:} Sentiment, NER, classification
    \item \textbf{FinancialQuantAgent:} Credit scoring, bankruptcy prediction, stock movement
\end{itemize}

Each specialist:
\begin{itemize}
    \item Cannot delegate further
    \item Uses restricted tools
    \item Must produce concise \texttt{<final\_answer>}
\end{itemize}

\subsection{Design Rationale}

The XML-based structure attempts to reduce output format drift, tool hallucination, nested JSON errors, and infinite reasoning loops.

In MAS, the Supervisor handles coordination only, while the Specialists handle domain execution. This separation enables controlled comparison of architectural effects under identical tooling constraints.

Both systems use identical tool definitions, follow identical structured reasoning format, and use the same base models. The only difference is reasoning decomposition (monolithic vs delegated), isolating the architectural effect.

\end{document}